\documentclass{article} 
\usepackage{iclr2025_conference,times}
\iclrfinalcopy

\usepackage{amsmath,amsfonts,bm}

\def\eqref#1{equation~\ref{#1}}

\def\1{\bm{1}}

\DeclareMathAlphabet{\mathsfit}{\encodingdefault}{\sfdefault}{m}{sl}
\SetMathAlphabet{\mathsfit}{bold}{\encodingdefault}{\sfdefault}{bx}{n}

\newcommand{\KL}{D_{\mathrm{KL}}}

\usepackage[utf8]{inputenc} 
\usepackage[T1]{fontenc}    
\usepackage{hyperref}       
\usepackage{url}            
\usepackage{booktabs}       
\usepackage{amsfonts}       
\usepackage{nicefrac}       
\usepackage{microtype}      
\usepackage{xcolor}         
\usepackage{dsfont}
\usepackage{tikz} 
\usepackage{mathtools}
\usetikzlibrary{shapes,decorations,arrows,calc,arrows.meta,fit,positioning}
\tikzset{
    -Latex,auto,node distance =1 cm and 1 cm,semithick,
    state/.style ={ellipse, draw, minimum width = 0.7 cm},
    point/.style = {circle, draw, inner sep=0.04cm,fill,node contents={}},
    bidirected/.style={Latex-Latex,dashed},
    el/.style = {inner sep=2pt, align=left, sloped}
}

\DeclarePairedDelimiterX{\infdivx}[2]{(}{)}{%
  #1\;\delimsize\|\;#2%
}

\newcommand{\bigCI}{\mathrel{\text{\scalebox{1.07}{$\perp\mkern-10mu\perp$}}}}

\title{Meta Learning not to Learn: Robustly Informing Meta Learning under Nuisance-Varying Families}

\author{%
  Louis McConnell \\
  Biomedical Data Science Center \\
  Lausanne University Hospital \\
  Lausanne, CH  \\
  \texttt{louie.mc@berkeley.edu} \\
}

\begin{document}

\maketitle

\begin{abstract}
In settings where both spurious and causal predictors are available, standard neural networks trained under the objective of empirical risk minimization (ERM) with no additional inductive biases tend to have a dependence on a spurious feature. As a result, it is necessary to integrate additional inductive biases in order to guide the network toward generalizable hypotheses. Often these spurious features are shared across related tasks, such as estimating disease prognoses from image scans coming from different hospitals, making the challenge of generalization more difficult. In these settings, it is important that methods are able to integrate the proper inductive biases to generalize across both nuisance-varying families as well as task families. Motivated by this setting, we present \textbf{RIME} (\textbf{R}obustly \textbf{I}nformed \textbf{M}eta l\textbf{E}arning), a new method for meta learning under the presence of both positive and negative inductive biases (what to learn and what not to learn). We first develop a theoretical causal framework showing why existing approaches at knowledge integration can lead to worse performance on distributionally robust objectives. We then show that RIME is able to simultaneously integrate both biases, reaching state of the art performance under distributionally robust objectives in informed meta-learning settings under nuisance-varying families\footnote{\href{https://github.com/louieg888/inp_discovery}{Code available here.}}. 
\end{abstract}

\section{Introduction}

Consider one of the central problems of radiology: identifying disease progression from medical image data. It is often the case that both the disease status and the image are associated with environment-dependent predictive factors such as the presence of medical devices that are predictive of the site chosen. Explainability studies of COVID-19 X-ray scan models have shown significant activation outside of the lung region
(\citet{antony_challenges_2023}), and age has been shown as a common nuisance in the prediction of disease prognosis (\citet{zhang_metapred_2019}, \citet{sanderman_age_2006}). 

Identifying solutions for these issues is a challenging technical problem due to the heterogeneity of data, tasks, and environments. Although it is possible to filter by differences such as site and disease, this approach does not leverage the common patterns and structure shared across  tasks and suffers from an inability to generalize to low data settings such as rare disease progression estimation.

These problem settings are dominated by two different types of heterogeneities. The first type of heterogeneity results from spurious relationships in which the distribution between a nuisance feature and the causal factor changes in different environments. The second type of heterogeneity is task variability, in which the relationship between causal factors and our outcome is different in the aggregated dataset (such as in the case of predicting the statuses of different diseases). We should leverage approaches that consider each of these complications in order to model heterogeneous objectives and environments. 

Environmental variability can be addressed by methods from the \hyperref[ood]{out of distribtion (OOD)} literature such as aggregation of risks from multiple environments or learning environment-invariant representations. Task variability can be addressed by approaches from \hyperref[meta_learning]{meta learning}, a field of research seeking to identify learning frameworks that optimize over the space of strategies for training machine learning models. In order to model both variabilities, a hybrid approach capable of integrating inductive biases about environments into a meta learning framework is needed.



Integrating prior knowledge in a meta learning context has been explored by \citet{kobalczyk_informed_2024} \footnote{Connections between informing of prior knowledge and the OOD literature can be found \hyperref[ood]{here}.}. They show that positive inductive biases (i.e. functional parameters) can inform the posterior over hypothesis classes and improve the sample efficiency of meta learning approaches. However, when negative inductive biases (i.e. invariances) constrain the space of functions with information that is not in the training data in the infinite data regime, improvements in sample efficiency can come from increased dependence on a spurious variable, resulting in worse OOD performance.

In this work, we investigate this setting of meta learning over a nuisance-varying family of tasks in the presence of positive and negative inductive biases. Our contributions are as follows: 1) We provide a causal formalism of the data generating process for robustly informed meta learning, explicitly outlining the assumptions of our method and explaining why positively informed meta learning frameworks are not only insufficient but worsen OOD performance under nuisance-varying families. 2) Under this setting, we present RIME, a method for integrating both positive and negative biases simultaneously. We establish a new synthetic benchmark for the objective of informed meta learning under nuisance-varying task families. We then test our method on this benchmark under combinations of positive and negative inductive biases in a nuisance-varying family and show its effectiveness over both traditional/informed meta learning under distributionally robust risk objectives.

\section{Background: Problem Formulation and Motivation}

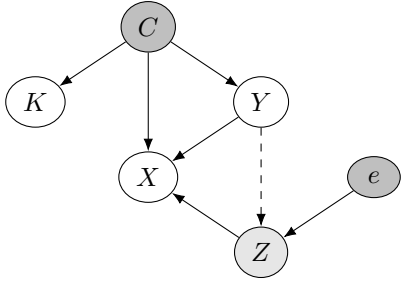
\begin{figure}\label{fig1}
  \centering
    \begin{tikzpicture}
        \node[state][fill=gray!50] (c) at (1.5,0) {$C$};
        \node[state] (k) at (0,-1) {$K$};
        \node[state] (y) at (3,-1) {$Y$};
        \node[state] (x) at (1.5, -2) {$X$};
        \node[state][fill=gray!20] (z) at (3, -3) {$Z$};
        \node[state][fill=gray!50] (e) at (4.5, -2) {$e$};

        \path (c) edge (k);
        \path (c) edge (x);
        \path (c) edge (y);
        \path (y) edge (x);
        \path (y) edge [dashed] (z);
        \path (z) edge (x);
        \path (e) edge (z);

    \end{tikzpicture}
  \caption{Causal structure of informed meta-learning under nuisance-varying families. Dark gray nodes represent latent variables, white nodes are observed variables, and light gray nodes are available only at training time. Dashed lines represent unstable (environment specific) relationships.}
\end{figure}
In this section we formally define the problem setting for RIME. This will lay the groundwork for our method and provide some theoretical insight into why vectorized knowledge representations cannot address negative inductive biases and can worsen OOD performance.

We are interested in learning over a distribution of related tasks $\{T \sim p(T) \}$, environments $\{e \sim p(e)\}$, and knowledge representations $\{K = f(T)\}$, where $f: \mathbb{R}^c \rightarrow \mathbb{R}^k$ is some deterministic knowledge extraction function of the task\footnote{We will use the symbol $T$ to represent the task itself and $C$ to represent the latent representation of the task.}. Recall that there are two kinds of variability we are interested in: task variability and environment variability. We capture task variability in our causal model by a latent representation of the task $C \in \mathbb{R}^c$ which encodes all information about the task; similarly, we capture environmental variability in our model by a latent representation of the environment $e \in \mathcal{E}$ which encodes all information about the environment. 
Each task induces a distribution $p(x | y, z, C) p ( y | C )$, and each environment induces a distribution over the nuisance variable $p(z | y, e)$. The data in each task $\{(x_i, y_i)\}_{i=1}^{n}$ will be divided into two sets: a context set $\{(x_i, y_i)\}_{i=1}^{m}$ and a target set $\{(x_i, y_i)\}_{i=m+1}^{n}$. Our goal will be to develop an estimator that is able to predict the target factors $y_t$ of target outcomes $x_t$ from a small set of context points $\{(x_i, y_i)\}_{i=1}^{m}$ and a knowledge representations $k_t$ in unseen tasks and unknown environment. 

In addressing this problem we will make some key assumptions. Although we assume that we will see a diversity of tasks during the meta-learning process, we do not assume that we will see a diversity of environments. We also do not assume that we will have the nuisance variable at test time. Finally, we assume statistical independence between the task and the environment. 

The causal graph in \hyperref[fig1]{Figure 1} details our specific assumptions about the data generating process and the influence of task and environmental variability on the joint distribution. Causal factors ($y$) are generated under a task-specific prior distribution, $p(y | C)$. Causal factors have an environment dependent relationship with the nuisance variable $z$, inducing an environment-specific nuisance distribution $p(z | y, e)$. Both $z$ and $y$ have a stable causal effect on the outcome variable $x \sim p(x | y, z, C)$; that is, $x \bigCI e | y, z$. We assume that the knowledge $k$ we observe is an environment-independent projection of our task representation. We also assume that $C$ has a causal effect on the task-specific causal distribution $p(y | C)$ and the task-specific outcome distribution $p(y | x, C)$. 


We define $\mathcal{F}_e = \{p_e := p(\cdot | e)\}$ to be the nuisance-varying family that we seek to optimize over. The RIME objective is to learn a hypothesis $\hat{p}_{\theta}(y|x, C)$ to minimize the maximum KL divergence between $\hat{p}_{\theta}(y|x, C)$ and the true conditional $p(y| x, C)$ in any environment in the nuisance-varying family. Formally, we define the following risk objective: 

\begin{align}
R(\hat{p}) = \sup \limits_{e \in \mathcal{E}} - \mathbb{E}_{p(C)} \mathbb{E}_{p_e (x | C)}\displaystyle \KL (p_e(y | x, C) \Vert \hat{p}(y | x, C ) )
\end{align}

Under this causal data generating structure and set of assumptions, \textbf{knowledge about the context can worsen performance in OOD environments}. This is because the observation of $x$ during the inference procedure induces a conditional dependence between the knowledge $k$ and the spurious variable $z$, resulting in a lower-variance estimate of the spurious variable $z$. As a result, vectorized knowledge representations as in \citet{kobalczyk_informed_2024} are insufficient to minimize risk under nuisance-varying families for two reasons. First, although knowledge representations are able to provide a better posterior over the context variables $p(C | k)$, they also increase the weight that the model $p(y | x, C)$ has on the nuisance variable $z$ compared to $p(y | x)$, leading to worse performance on distributionally robust objectives than in the uninformed setting. In addition, even under fixed $C$, knowledge $k$ is not able to adjust estimates of $p(y|x, C)$ to mitigate the environment-specific mediating path between $x$ and $y$ through the nuisance $z$. This effect is \hyperref[section.4.2]{verified experimentally}.

\section{Robustly Informed Meta Learning (RIME)} \label{section.3.1}

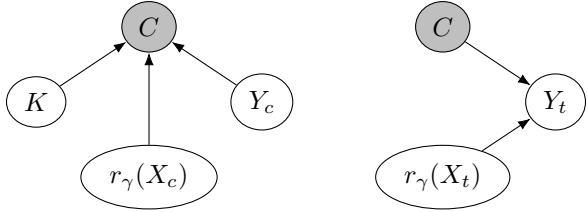
\begin{figure}\label{fig2}
    \centering
    \begin{tikzpicture}
        \node[state][fill=gray!50] (c) at (1.5,0) {$C$};
        \node[state] (k) at (0,-1) {$K$};
        \node[state] (y) at (3,-1) {$Y_c$};
        \node[state] (x) at (1.5, -2) {$r_{\gamma}(X_c)$};
    
        \path (x) edge (c);
        \path (y) edge (c);
        \path (k) edge (c);
    \end{tikzpicture}
    \hspace{1cm}%
    \begin{tikzpicture}
        \node[state][fill=gray!50] (c) at (1.5,0) {$C$};
        \node[state] (y) at (3,-1) {$Y_t$};
        \node[state] (x) at (1.5, -2) {$r_{\gamma}(X_t)$};
    
        \path (c) edge (y);
        \path (x) edge (y);
    \end{tikzpicture}    
    \caption{Causal structures for the encoding and decoding processes for RIME. In the encoding stage, information from the context set $(r_{\gamma}(x_c), y_c)$ and the prior knowledge representation $K$ is mapped to the context, a latent representation for the task function $f$. During decoding, the underlying causal factor $y_t$ is probabilistically inferred from the context and target $x$ variables $r_{\gamma}(x_t)$.}
    \label{fig:enter-label}
\end{figure}

We now present our main contribution, Robustly Informed Meta Learning (RIME), by building off of previous work in \citet{kobalczyk_informed_2024} and \citet{puli_out--distribution_2023} to meta-learn under nuisance-varying families and prior knowledge. To do so, we extend the work of \citet{puli_out--distribution_2023} to the informed meta-learning setting induced under our causal structure. Details about this work in the context of RIME are available in \hyperref[nurd_review]{Appendix A.4} and provide important background for this section. Informally, RIME seeks to isolate $z$ from the rest of the causal DAG by (1) applying inverse probability weighting to break the relationship between $y$ and $z$ and (2) removing the influence of $z$ on $x$ so that the nuisance does not affect the prediction of $y$. 

Under our causal structure, we have the following estimation of the nuisance-randomized density:

\begin{align}
p_{\bigCI}(x, y, z, C, k) = \frac{p_e(y)}{p_e(y | z)} \cdot p(x, y, z, C, k, e) = p(y | C)p_e(z) p(x | y, z, C) p (k | C) p (C)
\end{align}

which follows from the assumption of the posterior nuisance distribution being context independent, $z \bigCI C | y$. This task-independent nuisance assumption allows us to apply reweighting across tasks without estimating task-specific weights for all contexts. 

To control for the influence of our nuisance in $x$, we enforce that $z$ provides no information beyond what is inferred from the context and the representation of $x$, or that $y \bigCI_{p_{\bigCI}} z | (r_{\gamma}(x), C)$. As in the single-task case in \citet{puli_out--distribution_2023}, we would also like to enforce that no information about spurious representations are encoded in the data ($z \bigCI_{p_{\bigCI}} r_{\gamma}(x)$) to avoid representations of $x$ that satisfy the conditional independence by encoding all information about the nuisance $z$. These two conditions are met if   $\mathbb{I}_{p_{\bigCI}}[(C, r_{\gamma}(x), y); z]$ is 0 \hyperref[proofs]{(Lemma 1)}. We can decompose this loss into the following components: 

\begin{align}
    \mathbb{I}_{p_{\bigCI}}[(C, r_{\gamma}(x), y); z] = \mathbb{I}_{p_{\bigCI}}[C; z | r_{\gamma}(x), y] + \mathbb{I}_{p_{\bigCI}}[(y, r_{\gamma}(x)); z]
\end{align}

The second term in this expression is equal to the single-task loss from \citet{puli_out--distribution_2023}. The first term is unique to the RIME setting and is 0 if we have the case that distillation perfectly enforces the independence relation $z \bigCI_{p_{\bigCI}} r_{\gamma}(x)$ and $y \bigCI_{p_{\bigCI}} z | r_{\gamma}(x)$, as these two constraints block all paths between $z$ and $C$ in our DAG over $p_{\bigCI}$. However, if the distillation condition is not fully met ($z$ is not fully independent of some $\hat{r}_{\gamma}(x)$), then conditioning on $\hat{r}_{\gamma}(x)$ will still induce a conditional dependence between the spurious variable $z$ and $C$ because $\hat{r}_{\gamma}(x)$ remains a collider in the causal graph. As this distillation condition will not be perfectly satisfied in practical settings, we minimize the mutual information loss $\mathbb{I}_{p_{\bigCI}}[(C, \hat{r}_{\gamma}(x), y); z]$ to control for the first term as well. Details about this problem and experimental validation of this effect are provided in \hyperref[section.4.2.1]{Section 4.2.1}. 


We formulate the preliminary objective is as follows: 

\begin{align}
\max \limits_{\theta, \gamma} \mathbb{E}_{\hat{p}_{\bigCI}(x, y, z, C, k)} \log p_{\theta}(y_t | r_{\gamma}(x_t), r_{\gamma}(x_c), y_c, k) - \lambda \mathbb{I}_{\hat{p}_{\bigCI}} [k, y_t, y_c, r_{\gamma}(x_t), r_{\gamma}(x_c); (z_t, z_c)]
\end{align}

In order to make our estimator $p_{\theta}$ suitable for meta learning over this objective, we parameterize $p_{\theta}$ as an Informed Neural Process (\citet{kobalczyk_informed_2024}). This method presents a natural way to perform posterior inference over the latent context distribution $p(C | x_c, y_c, k)$ and sample from an evaluatable functional space, making it a natural fit for knowledge integration in meta learning. The ELBO for their approach under our parameterized representation is as follows: 

\begin{align}
    \log p(y_t | r_{\gamma}(x_t), r_{\gamma}(x_c), y_c, k) \geq \mathbb{E}_{q(C | r_{\gamma}(x_c), y_c, k)} [\log p(y_t | r_{\gamma}(x_t), C)] \\
    - \KL (q(C | r_{\gamma}(x_t), y_t, k) \Vert q(C | r_{\gamma}(x_c), y_c, k)
\end{align}

Combining these two objectives, we have the final RIME objective:

\begin{align}
    \mathcal{L}_1 &= -\mathbb{E}_{q(C | r_{\gamma}(x_c), y_c, k)} [\log p(y_t | r_{\gamma}(x_t), C)] \\
    \mathcal{L}_2 &= \KL (q(C | r_{\gamma}(x_t), y_t, k) \Vert q(C | r_{\gamma}(x_c), y_c, k) \\
    \mathcal{L}_3 &= \mathbb{I}_{\hat{p}_{\bigCI}} [k, y_t, y_c, r_{\gamma}(x_t), r_{\gamma}(x_c); (z_t, z_c)] \\
    \mathcal{L}_{RIME} &= \mathcal{L}_1 + \beta \mathcal{L}_2 + \lambda \mathcal{L}_3
\end{align}

for scalar hyperparameter coefficients $\beta, \lambda$.

\hyperref[fig2]{Figure 2} illustrates the data flow for inference as well as training of $\theta, \gamma$ in the RIME method. During the encoding stage, the task knowledge, outcome representation context variables $r_{\gamma}(x_c)$, and causal factor context variables $y_c$ are encoded into a latent representation of the task, $C$. During the decoding stage, the context $C$ and outcome representation target variables $r_{\gamma}(x_t)$ are decoded and mapped to a distribution over $y_t$. Details for training RIME regarding specific implementations of mutual information enforcement and inverse probability weighting can be found in \hyperref[training_details]{Appendix C}.

\section{Experiments}

\subsection{Experiment 1: No Task Variability}

In both experiments, we adopt the setting in \citet{puli_out--distribution_2023} and construct nuisance-varying class conditional Gaussians. The nuisance and causal factor are mixed with varying amounts of noise in the outcomes $x$, making it easier to learn a biased shortcut via the lower-noise mixture. Our hypothesis is that RIME will remove this shortcut and result in higher OOD performance.

In the first experiment, we validate that neural reweighing and distillation are able to improve existing informed meta-learning approaches in the single task nuisance-varying family setting. These variables are distributed over nuisance-varying family $\mathcal{F}_e$ as follows: 

\begin{align}
y \sim \mathcal{B}(0.5), z \sim \mathcal{N}(e(2y-1)), x = [x_1 \sim \mathcal{N}(3y-z, 9), x_2 \sim \mathcal{N}(3y+z, 0.01)]
\end{align}

where the optimal decorrelating representation is $x^* = x_1 + x_2$. In this experiment, we tested 3 setups. The first is a vanilla neural process trained with $x$ and $y$. The second is an architecture in which the optimal representation for this problem was fixed. In the final setup, we use a discriminator model to estimate the mutual information and learned the representation during training. 

We see a dramatic improvement in the OOD performance of both the learned and the optimal representation compared to the vanilla neural process. We observe a small difference in performance between the ground truth optimal representation and the learned representation, with the non-optimal representation performing slightly better in-distribution but equally well out of distribution. Data and figures for this experiment, as well as additional experimental details, are available in \hyperref[experimental_results_plots]{Appendix D})

\subsection{Experiment 2: Task Variability}\label{section.4.2}

\begin{figure}[h]\label{fig3}
    \centering
    \includegraphics[width=\textwidth]{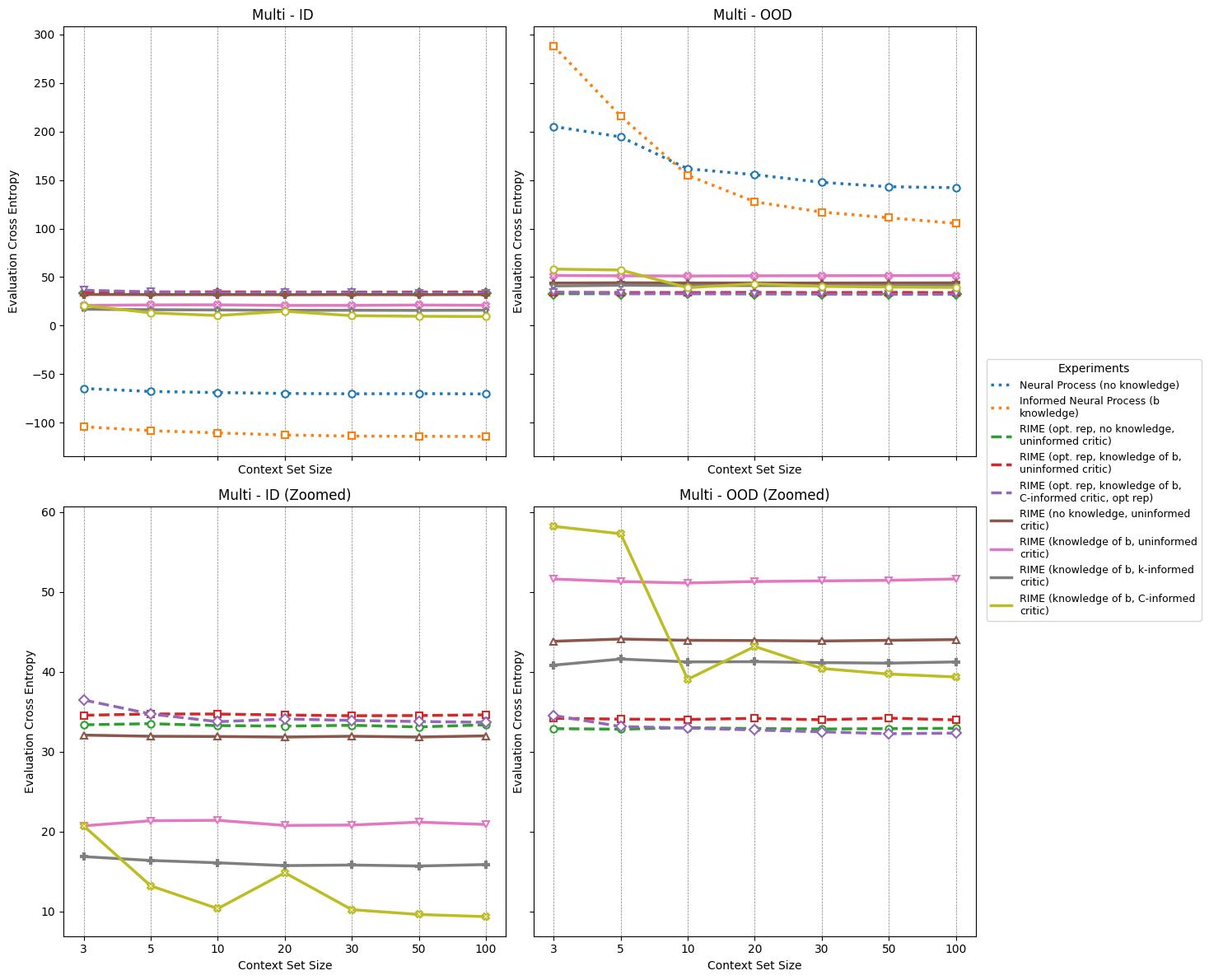}  
    \caption{$k$-shot evaluation cross entropy loss from experiment 2 (lower is better; risk evaluated by lowest line). Dotted lines do not use reweighting / distillation, solid lines do. Bottom plot is a zoomed-in plot of different RIME variants and illustrates the effect of knowledge integration / critic setups as well as experiments using the optimal representation as baselines. Best-performing methods with learned representations are in grey ($k$-informed critic, knowledge of $b$) and lime ($C$-informed critic, knowledge of $b$).}
\end{figure}

The second experiment extends the first experiment to a meta learning setting by adding in an additional knowledge parameter and shows the effectiveness of RIME in integrating additional knowledge, both in distribution and out of distribution.  We have the following setting:

\begin{align}
y \sim \mathcal{B}(0.5), z \sim \mathcal{N}(e(2y-1)), x = [x_1 \sim \mathcal{N}(b + 3y - z, 9), x_2 \sim \mathcal{N}(b + 3y + z, 0.01)]
\end{align}

where $b \sim \mathcal{U}[-2, 2]$. We find that without RIME, although the knowledge helps the performance in distribution, it severely worsens the performance in the few-shot out of distribution setting. However, we also find that \textbf{under RIME (knowledge-informed critic), positive knowledge improves performance both in distribution and out of distribution relative to the uninformed setting, at every $k$-shot metric}. This shows that RIME is able to improve the quality of integration of causal knowledge about the task under the presence of nuisance-varying families. Data, figures, and details for this experiment are available in \hyperref[experimental_results_plots]{Appendix D}.

In the following sections, we will dive more deeply into two key observations from the experiments.

\subsubsection{Informed Critics Improve Robustness under Noisy $I_{p_{\bigCI}}$ Enforcement}\label{section.4.2.1}

As discussed in \hyperref[section.3.1]{Section 3}, a task-independent critic is theoretically valid given perfect enforcement of our mutual information criteria. In practice we find that due to errors in both critic loss and mutual information enforcement, the distilled representation retains some mixing of $z$ in the final representation. We hypothesize that in the multitask setting, this results in knowledge being leveraged to identify a shortcut through the spurious feature and consequently worse generalization. We can see this effect in \hyperref[fig3]{Figure 3} in the zoomed plot. In the setting in which we use an uninformed critic in RIME, we find that knowledge increases in distribution performance but decreases out of distribution performance. However, this ID-OOD gap is no longer present when we set $r_{\gamma}(x)$ to the optimal representation, indicating that an imperfect $r_{\gamma}(x)$ is the cause of poor knowledge integration over our risk objective. Adding the additional constraint of $I_{p_{\bigCI}}[C; z | y, r_{\gamma}(x)]$ improves the quality of our representation, leading to effective knowledge integration both in distribution and out of distribution.



\subsubsection{RIME Increases Multitask Sample Efficiency under Spurious Correlations}

In the zoomed out plots in \hyperref[fig3]{Figure 3}, we can see that in the OOD setting, models that do not use RIME see performance increases as a function of context set size. However, under RIME, the sample efficiency is significantly higher with no significant performance differences between the 3-shot and 100-shot settings except with the context-informed critic. We hypothesize that this dependence is due to the fact that the low-shot regime compromises the quality of context representation, resulting in higher variance estimates in low-shot regimes. However, in the knowledge-informed critic setting, this instability is not present, resulting in superior sample efficiency.

\section{Discussion, Limitations and Future Work}


Neural Processes frequently underfit the data provided, suggesting that the architecture is not well suited for data with high intrinsic dimension or noisy samples. Attention-based mixing functions suggested in \citet{kim2019attentiveneuralprocesses} and the noise-robust loss suggested in \citet{shapira_robust_2024} could potentially alleviate this problem and would be an interesting addition. Adapting this particular model to real world data with complex relationships would be a valuable extension. In our experiments, we chose settings in which there was an optimal representation. It would be interesting to characterize the conditions under which this optimal representation exists and characterize the evaluation tradeoffs that we see as a function of the significance of the nuisance on the outcome. 


\bibliography{iclr2025_conference}
\bibliographystyle{iclr2025_conference}

\appendix

\section{Related Work}

\subsection{Meta Learning}\label{meta_learning}

The key objective in meta learning is to optimize the learning process over a set of learning tasks. The tasks are divided into training/validation/test partitions, and meta learning models are evaluated by how efficiently they are able to learn new tasks. For some input $x$, output $y$, and task $C$ distributed under $p(x, y, C)$, the objective in meta learning is to maximize the likelihood $p_{\theta}(y | x, C)$ over the joint distribution $p(x, y, C)$. In optimizing this objective, there are three primary classes of meta learning methods used. 

\textbf{Metric-based Approaches.} These approaches in general try to learn a distance metric $k_\theta(x, x')$ between any two input points in the dataset and then use this distance to weigh the labels of the support set. \citet{koch_siamese_nodate} uses Siamese networks to learn embeddings and use a kernel to do one shot per class image classification. \citet{snell_prototypical_2017} explicitly develops "prototype representations" of each class and performs classification by computing nearest neighbors over the prototypical representations. \citet{vinyals_matching_2017} explicitly learns a kernel $k(x, x')$ and performs prediction by weighing the labels of a support set by this kernel function. 

\textbf{Gradient-based Approaches.} The goal behind these methods is to discover an optimal initial set of parameters $\hat{\theta}$ for $p_{\theta}(y | x, C)$ such that all tasks only require a few gradient steps to have competitive performance (\citet{finn_model-agnostic_2017}, \citet{nichol_reptile_nodate}, \citet{li_meta-sgd_2017}). 

\textbf{Model-based Approaches.} These methods explicitly model $p_{\theta}(y | x, C) = f_{\theta}(x, C)$ via inference of $C$ and include methods such as Neural Processes (\citet{garnelo_neural_2018}, \citet{garnelo_conditional_2018}) which learn to build a latent representation of a function from a few context points in the task family and learn a process to sample from the task representation given a new input point. These types of  models are the most natural fit for RIME as they can consider additional knowledge via the context variable and explicitly model a representation of the task being learned, allowing for a rigorous theoretical analysis of the relationships between environment shifts and task shifts.  

\subsection{Informed Machine Learning}\label{informed_ml}

Informed machine learning is a field of machine learning research that seeks to integrate prior knowledge and inductive biases into machine learning algorithms. Prior knowledge is referred to as any knowledge representation outside of the dataset that limits the hypothesis class $f_{\theta}$. 

\citet{von_rueden_informed_2023} have a comprehensive taxonomy of informed machine learning and discusses in depth the various types of knowledge that can be integrated and the way that they have been integrated into existing machine learning methods, an overview of which will be summarized here. Prior works have integrated knowledge in a variety of different forms, including logic rules (\citet{8260755}, \citet{xu_semantic_2018}), algebraic equations (\citet{daw_physics-guided_2021}, \citet{stewart_label-free_2016}), differential equations (\citet{raissi2017physicsinformeddeeplearning}), knowledge graphs (\citet{marino_more_2017}, \citet{jiang2018hybridknowledgeroutedmodules}), object relational knowledge (\citet{battaglia2016interactionnetworkslearningobjects}), vector knowledge representations (\citet{kobalczyk_informed_2024}), and statistical / causal knowledge (\citet{pawlowski_deep_2020}). In the aforementioned papers the most common method of integration is via adding an auxiliary loss objective to the optimization function (\citet{8260755}, \citet{xu_semantic_2018}, \citet{daw_physics-guided_2021}, \citet{stewart_label-free_2016}). Other methods of integration vary by the type of knowledge being leveraged. In \citet{pawlowski_deep_2020}, causal knowledge is integrated by factorizing the representation of each node in a causal graph into a function of only its parents. The method for integration for knowledge graphs and object-relational graphs is more intricate and crafted into the underlying architecture of the method (\citet{marino_more_2017}, \citet{battaglia2016interactionnetworkslearningobjects}, \citet{jiang2018hybridknowledgeroutedmodules}). \citet{kobalczyk_informed_2024} presents the only work to the authors' knowledge concerning the integration of meta-learning and informed machine learning, but knowledge integration is limited to vector knowledge representations with no additional loss penalty. As we show, this will ultimately limit the degree to which this approach is able to integrate knowledge that targets improvements over distributionally robust objectives, such as causal or statistical knowledge. 

\subsection{Out of Distribution (OOD) Generalization, Spurious Correlations, and Connections to Informed ML} \label{ood}

Out of Distribution Generalization in the presence of spurious correlations is a problem that is intimately tied to informed machine learning. In both fields there is an underlying assumption that additional inductive biases will improve a machine learning method's performance. In the traditional supervised setting under ERM, it is not possible to do any better in shifted environments without additional inductive biases in the form of prior knowledge or a different risk objective (\citet{arjovsky_invariant_2020}, \citet{chen2023understandingimprovingfeaturelearning}). In settings with multiple predictive factors for a label and no further inductive biases, it has been shown by previous work that the two factors that influence which hypothesis a model learns in the hypothesis set under ERM are the inductive biases of the model class (\citet{atanov_task_2022}, \citet{battaglia2016interactionnetworkslearningobjects}) as well as the simplicity of learning the task from the predictor (\citet{shah_pitfalls_2020}). As a result, additional inductive biases are needed in order to improve performance (\citet{NEURIPS2021_efb76cff}, \citet{NEURIPS2021_c404a5ad}).

There are two high level approaches to providing inductive biases to achieve OOD generalization under spurious correlations. The first approach, Distributionally Robust Optimization (DRO), focuses on the modification of the statistical risk objective by considering a set of risks over multiple distributions that we are looking to optimize over (\citet{sagawa_distributionally_2020}, \citet{kuhn2024distributionallyrobustoptimization}, \citet{Rahimian_2022}, \citet{krueger2021outofdistributiongeneralizationriskextrapolation}). The second approach focuses on enforcing invariance over noncausal factors in order to provide the model with only causal predictors (\citet{arjovsky_invariant_2020}, \citet{kamath2021doesinvariantriskminimization}, \citet{nguyen2022domaininvariantrepresentationlearning}, \citet{chen2022learningcausallyinvariantrepresentations}). These invariances can be adversarially enforced (\citet{ganin_domain-adversarial_2016}), enforced through both traditional and counterfactual augmentations on the data (\citet{sachdeva2024catfoodcounterfactualaugmentedtraining}, \citet{deng-etal-2023-counterfactual}, \citet{1227801}), or enforced over the representation space of the input data (\citet{arjovsky_invariant_2020}, \citet{long2015learningtransferablefeaturesdeep}). The goal in both approaches is the same: either by changing the risk objective or by enforcing invariances, they seek to mitigate the degree to which a model weights spurious features by integrating prior knowledge of invariances or group labels. In this paper, we establish an approach that is able to combine the types of prior knowledge integration in informed machine learning and OOD generalization, reaching better performance than either method is capable of individually on distributionally robust statistical risks. 

\subsection{Nuisance Randomization and Uncorrelating Representations}\label{nurd_review}

In this section we explore one approach to dealing with the induced conditional dependence between $z$ and $C$ and the environment-specific mediating path in a simplified setting. \citet{puli_out--distribution_2023} addresses this simplified problem of learning under nuisance-varying families in our particular causal substructure given a fixed context and no knowledge integration in a traditional supervised setting. If there is no causal relationship between the nuisance and the outcome, it is sufficient to effectively break the nuisance-label relationship by performing inverse probability weighting. This enforces statistical independence in the reweighted joint distribution between the label and the nuisance, blocking the effect of purely spurious correlations. 

In this case, the joint probability distribution factorizes into the following: 

\begin{align}
p_{\bigCI}(x, y, z) = \frac{p_e(y)}{p_e(y | z)} \cdot p(x, y, z, e) = p(y)p_e(z) p(x | y, z) 
\end{align}

Weighting our original joint density by a factor of $\frac{p_e(y)}{p_e(y | z)}$ randomizes the relationship between $y$ and $z$, making them statistically independent. However, under these settings, two problems remain. Although this approach works well by itself in cases where there is no causal relationship between the nuisance and the outcome, estimators $p_{\bigCI}(y | x)$ can perform even worse than randomly sampling over the prior $p(y)$ in some contexts (\citet{puli_out--distribution_2023}). In these settings, the distribution over the nuisance and causal factor changes between environments, but the nuisance and label are environment-independently mixed in our outcome $x$. For example, age is a common nuisance for prognostic modeling, but it is often also the case that age is a biologically relevant feature as well (\citet{f93b76fd5b6143ad970ac08ff16b9e4c}). In these settings, optimizing over the adjusted joint distribution will still perform poorly under nuisance shifts due to the model effectively inferring the nuisance from the outcome variable and context and using it accordingly. 

In order to prevent our model from inferring $z$ to predict the label, we can learn an uncorrelating representation $r_{\gamma}(x)$ over the adjusted joint density such that $z$ no longer improves the prediction of $y$ from $r_{\gamma}(x)$ alone, formalized as $y \bigCI z | r_{\gamma}(x)$. This constraint is still not enough, as it is satisfied under functions that contain low-variance predictors of the nuisance, such as $f(x) = z$, as these functions do not improve the posterior of $y$ when conditioned on $z$ as well. To alleviate this problem we can also enforce $z \bigCI_{p_{\bigCI}} r_{\gamma}(x)$, yielding the joint objective of minimizing $\mathbb{I}_{p_{\bigCI}}[(y, r_{\gamma}(x)); z]$ by the chain rule. 

\section{Proofs}\label{proofs}

\subsection{Proof of Lemma 1}

By the chain rule of mutual information, we have 

\begin{align}
    \mathbb{I}_{p_{\bigCI}}[(C, r_{\gamma}(x), y); z] &= \mathbb{I}_{p_{\bigCI}}[y; z | (r_{\gamma}(x), C)] + \mathbb{I}_{p_{\bigCI}}[(r_{\gamma}(x), C); z]
\end{align}

The first term is 0 given the independence relation $y \bigCI_{p_{\bigCI}} z | (r_{\gamma}(x), C)$. The second term can be further decomposed as follows: 
\begin{align}
    \mathbb{I}_{p_{\bigCI}}[(r_{\gamma}(x), C); z] =  \mathbb{I}_{p_{\bigCI}}[(r_{\gamma}(x); z) | C] + \mathbb{I}_{p_{\bigCI}}[z; C] 
\end{align}

The first term is 0 by the independence relation $r_{\gamma}(x) \bigCI_{p_{\bigCI}} z$, which in the causal graph also implies $r_{\gamma}(x) \bigCI_{p_{\bigCI}} z | C$. The second term is 0 by $z \bigCI_{p_{\bigCI}} C$ implied in the causal graph under the balanced distribution with the induced statistical independence $z \bigCI_{p_{\bigCI}} y$. 








\section{Training Details for RIME}\label{training_details}

The training of RIME is divided into two stages. In the first stage of training, a probability weight is computed for each point in each training task. During the second stage of training, RIME learns the representation for outcomes $x$ and the context-dependent target estimator $p(y_t | r_{\gamma}(x_t), r_{\gamma}(x_c), y_c, k)$.

During the first stage, estimators for $p(y | C)$ as well as $p(z | y)$ are both learned by k-fold cross validation. The weights for each data point are computed from the validation sets to avoid overfitting. We parameterized these weighting models with an MLP.

In addition to learning the representations of $x$ and our target estimator, the second stage of training requires an estimation of the mutual information $\mathbb{I}_{p_{\bigCI}}[k, r_{\gamma}(x), y ; z]$. We use the density ratio trick for mutual information estimation as introduced in \citet{suzuki_approximating_2008} and train a discriminator $D_{\phi}(k, r_{\gamma}(x), y, z)$ on "fake" points sampled from the marginal density product $p(k, r_{\gamma}(x), y) p(z)$ and "real" points sampled from the joint density $p(k, r_{\gamma}(x), y, z)$. We can then estimate the mutual information as

\begin{align}
\mathbb{I}_{p_{\bigCI}}[k, r_{\gamma}(x), y ; z] = \mathbb{E}_{p_{\bigCI}(k, x, y, z)} \frac{D_{\phi}(k, x, y, z)}{1 - D_{\phi}(k, x, y, z)}
\end{align}

and optimize over this estimate while training $p_{\theta}$. 

We tested informing the critic with either (a) the knowledge $k$ or (b) the task latent representation $C$ in Experiment 2. We find that performance is more stable when conditioning on the knowledge representation, as seen in Figure 4. We hypothesize that the dynamic relationship between the context representation, outcome representation, and critic model may contribute to training instability. In spite of this, context-informed critics may outperform knowledge-informed critics under more complex functions where prior knowledge does not effectively capture the full latent context.

The second stage of training alternates between optimizing the discriminator model for mutual information estimation  and optimizing the representation / likelihood estimation model via a two-step process. In the first step, training is carried out episodically, with each task divided into target sets and context sets. Before the context sets are sampled from the target sets, the target sets are upsampled according to the probability weights computed in stage 1. A batch of training tasks is then sampled from the training distribution, the $x$ values are mapped to $r_{\gamma}(x)$, and the causal factors $y_t$ are inferred for the target and the loss over the $r_{\gamma}$ and $p_{\theta}$ is backpropagated. The relative frequency of each step is a hyperparameter (we did 8 critic updates per predictive update). 

For more details about the training procedure, see the code released \href{https://github.com/louieg888/inp_discovery}{here}.

\section{Complete Experimental Details, Results, and Plots}\label{experimental_results_plots}

\subsection{Additional Experimental Details}

In RIME experiments, we estimate the reweighing coefficients over the training set by 5-fold cross validation. Once these weights are computed the training set is upsampled by a factor of 10, and these upsampled points make up the target set. 

In all experiments, we used tasks of 100 target points each that were upsampled into 1000 target points. In the standard setting they are upsampled uniformly, while in the RIME setting they are upsampled by the estimated weights. We use an upsampling approach instead of a gradient-based approach as \citet{li2024upsampleupweightbalancedtraining} shows lower loss variance due to optimizational stochasticity. The context points are sampled from the target points; the number of context points is sampled uniformly on $\mathcal{U}[0, 100]$.

All experimental results are averaged over 10 seeds. For each seed, all metrics were computed on the evaluation step with the minimum in-distribution (ID) validation loss, as the out of distribution (OOD) environments will not be visible at evaluation in train or test data. In order to avoid biasing any particular k-shot metric in the step selection, the context size for in-distribution evaluation loss for step selection is a Monte-Carlo estimate of the average evaluation loss over the uniform distribution of all context sizes between the minimum and maximum context size.

\subsection{Tables and Plots}\

\begin{figure}[h]
    \centering
    \includegraphics[width=\textwidth]{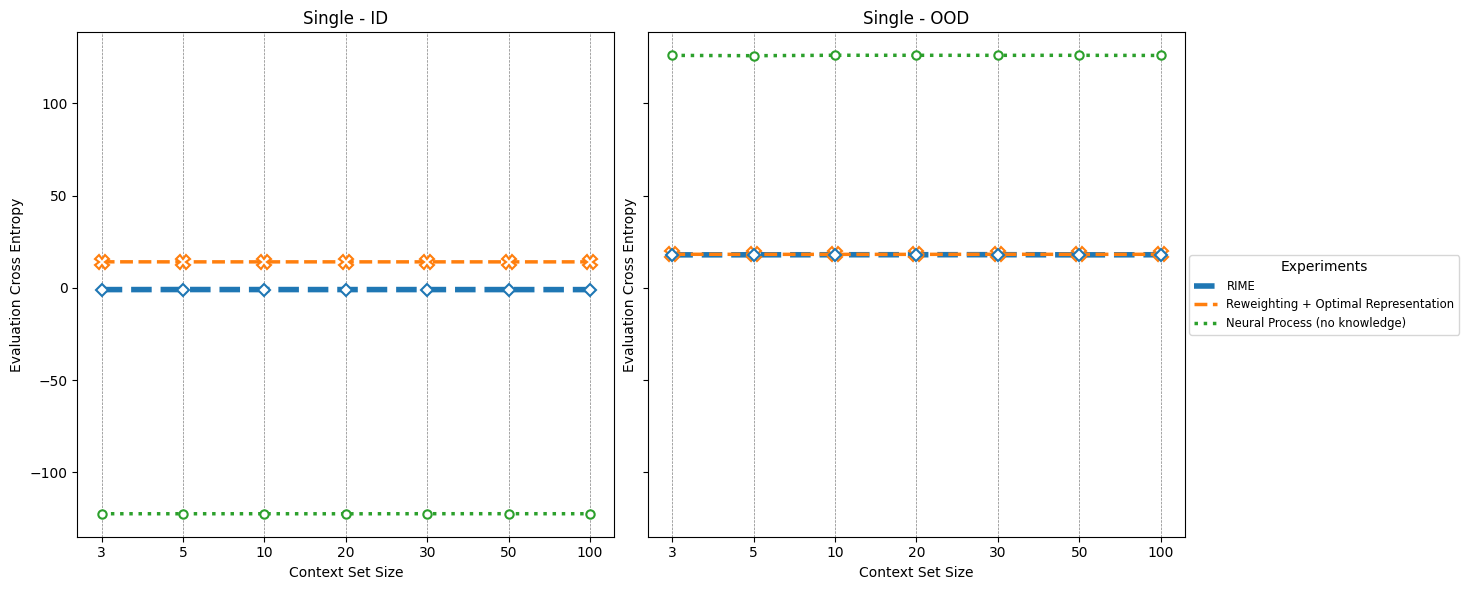}  
    \caption{K-shot evaluation cross entropy loss from experiment 1 (lower is better). Dotted lines do not use reweighting / distillation; dashed lines do.}
\end{figure}


\begin{table}[h]
  \caption{Target Cross Entropy Loss for Experiment 1: Single Task}
  \label{merged-results-table}
  \centering
  \begin{tabular}{lcccccc}
    \toprule
     & \multicolumn{6}{c}{\textbf{In-Distribution} $(e = 0.5)$} \\
    \cmidrule(lr){2-7}
    Context set size (\# shots) & 3 & 5 & 10 & 20 & 50 & 100 \\
    \midrule
    Neural Process                        & \textbf{-122} & \textbf{-122} & \textbf{-122}& \textbf{-122} & \textbf{-122} & \textbf{-122} \\
    \midrule
    RIME (opt. rep)   & 14.1 & 14.2 & 14.1 & 14.1 & 14.1 & 14.1 \\
    RIME                                  & -0.905 & -0.871 & -0.890 & -0.892 & -0.898 & -0.910 \\
    \midrule
     & \multicolumn{6}{c}{\textbf{Out-of-Distribution} $(e = -0.9)$} \\
    \cmidrule(lr){2-7}
    Context set size (\# shots) & 3 & 5 & 10 & 20 & 50 & 100 \\
    \midrule
    Neural Process                  & 126 & 126 & 126 & 126 & 126 & 126 \\
    \midrule
    RIME (opt. rep)  & 18.2 & 18.2 & 18.2 & 18.2 & 18.2 & 18.2  \\
    RIME                            & \textbf{17.9} & \textbf{18.0} & \textbf{18.0} & \textbf{18.0} & \textbf{17.9} & \textbf{18.0} \\
    \bottomrule
  \end{tabular}
\end{table}

\begin{table}[h]
  \caption{Target Cross Entropy Loss for Experiment 2: Multitask}
  \label{merged-results-table}
  \centering
  \begin{tabular}{lcccccc}
    \toprule
     & \multicolumn{6}{c}{\textbf{In-Distribution} $(e = 0.5)$} \\
    \cmidrule(lr){2-7}
    Context set size (\# shots) & 3 & 5 & 10 & 20 & 50 & 100 \\
    \midrule
    Neural Process (no knowledge)                      & -64.8 & -67.9 & -69.0 & -69.9 & -70.1 & -70.4 \\
    Informed Neural Process (knowledge of $b$)         & \textbf{-104} & \textbf{-108} & \textbf{-110} & \textbf{-113} & \textbf{-114} & \textbf{-114} \\
    \midrule
    RIME (opt. rep, no knowledge, uninformed critic)    & 33.4 & 33.5 & 33.3 & 33.2 & 33.3 & 33.4 \\
    RIME (opt. rep, knowledge of $b$, uninformed critic)    & 34.5 & 34.7 & 34,7 & 34.6 & 34.5 & 34.6 \\
    RIME (opt. rep, knowledge of $b$, c-informed critic)    & 36.5 & 34.7 & 33.7 & 34.1 & 33.9 & 33.7 \\
    \midrule
    RIME (no knowledge, uninformed critic)                       & 32.1 & 31.9 & 31.9 & 31.8 & 31.8 & 32.9 \\
    RIME (knowledge of $b$, uninformed critic)                      & 20.7 & 21.3 & 21.4 & 20.8 & 21.2 & 20.9 \\
    RIME (knowledge of $b$, k-informed critic)    & 16.9 & 16.4 & 16.1 & 16.7 & 15.7 & 15.9 \\
    RIME (knowledge of $b$, c-informed critic)    & 20.7 & 13.2 & 10.3 & 14.8 & 10.2 & 9.3 \\

    \midrule
     & \multicolumn{6}{c}{\textbf{Out-of-Distribution} $(e = -0.9)$} \\
    \cmidrule(lr){2-7}
    Context set size (\# shots) & 3 & 5 & 10 & 20 & 50 & 100 \\
    \midrule
    Neural Process (no knowledge)                      & 205 & 194 & 162 & 155 & 148 & 143 \\
    Informed Neural Process (knowledge of $b$)         & 288 & 216 & 154 & 128 & 111 & 105 \\
    \midrule
    RIME (opt. rep, no knowledge, uninformed critic)    & 32.9 & 32.8 & 33.0 & 32.9 & 32.8 & 32.9 \\
    RIME (opt. rep, knowledge of $b$, uninformed critic)    & 34.2 & 34.1 & 34.0 & 34.2 & 34.0 & 34.0 \\
    RIME (opt. rep, knowledge of $b$, $C$-informed critic)    & 34.5 & 33.1 & 32.9 & 32.7 & 32.5 & 32.3 \\
    \midrule
    RIME (no knowledge, uninformed critic)                       & 43.8 & 44.1 & 43.9 & 43.9 & 43.8 & 43.9 \\
    RIME (knowledge of $b$, uninformed critic)                      & 51.5 & 51.3 & 51.1 & 51.3 & 51.4 & 51.6 \\
    RIME (knowledge of $b$, $k$-informed critic)    & \textbf{40.8} & \textbf{41.6} & 41.2 & \textbf{41.3} & 41.1 & 41.2 \\
    RIME (knowledge of $b$, $C$-informed critic)    & 58.2 & 57.3 & \textbf{39.0} & 43.2 & \textbf{40.4} & \textbf{39.3} \\
    \bottomrule
  \end{tabular}
\end{table}

\section{Acknowledgements}

Thanks to Romain Lopez, Claudia Iriondo, Spencer Song, and Dylan Doblar for helpful comments, insights, editing, and feedback. 

\end{document}